%% file: acl2019.tex
\documentclass[11pt,a4paper]{article}
\usepackage[hyperref]{emnlp-ijcnlp2019}
\usepackage{times}
\usepackage{latexsym}

\input{math_commands.tex}

\usepackage{hyperref}
\usepackage{url}

\usepackage{amsmath}
\usepackage{amssymb}
\usepackage{graphicx}
\usepackage{multirow}
\usepackage{subfig}
\usepackage{url}
\usepackage{paralist}
\usepackage{latexsym}
\usepackage{amsmath}
\usepackage{color}
\usepackage{graphicx}
\usepackage{booktabs}
\usepackage{subfig}
\usepackage{url}
\usepackage{tabularx}
\usepackage{multirow}
\usepackage{verbatim}
\usepackage{algorithm}
\usepackage{algpseudocode}
\usepackage{CJK}

\aclfinalcopy 

\title{Dynamic Past and Future for Neural Machine Translation}

\author{
	Zaixiang Zheng \\ Nanjing University \\ {\normalsize \tt zhengzx@smail.nju.edu.cn} \\\And
	Shujian Huang \\ Nanjing University  \\ {\normalsize \tt huangsj@nju.edu.cn} \\\And
	Zhaopeng Tu \\ Tencent AI Lab \\ {\normalsize \tt zptu@tencent.com} \\\AND
	Xin-Yu Dai \\ Nanjing University \\   {\normalsize \tt dxy@nju.edu.cn} \\\And
	Jiajun Chen \\ Nanjing University \\ {\normalsize \tt chenjj@nju.edu.cn}
}

\date{}

\def\past{\textsc{Past}}
\def\future{\textsc{Future}}
\def\present{\textsc{Present}}

\begin{document}

\maketitle
\begin{abstract}
Previous studies have shown that neural machine translation (NMT) models can benefit from explicitly modeling translated (\past) and untranslated (\future) source contents as recurrent states~\cite{zheng2018modeling}. 
However, this less interpretable recurrent process hinders its power to model the dynamic updating of \past\ and \future\ contents during decoding.
In this paper, we propose to model the \textit{dynamic principles} by explicitly separating source words into groups of translated and untranslated contents through parts-to-wholes assignment.
The assignment is learned through a novel variant of routing-by-agreement mechanism~\cite{sabour2017dynamic}, namely {\em Guided Dynamic Routing}, where the translating status at each decoding step \textit{guides} the routing process to assign each source word to its associated group (i.e., translated or untranslated content)  represented by a capsule, enabling translation to be made from holistic context.
Experiments show that our approach achieves substantial improvements over both \textsc{Rnmt} and Transformer by producing more adequate translations. Extensive analysis demonstrates that our method is highly interpretable, which is able to recognize the translated and untranslated contents as expected.\footnote{Codes are released at \url{https://github.com/zhengzx-nlp/dynamic-nmt}.}
\end{abstract}

\section{Introduction}
\label{sec:intro}

Neural machine translation~(NMT) generally adopts an \textit{attentive encoder-decoder} framework~\cite{sutskever2014sequence,Vaswani2017Attention}, where the encoder maps a source sentence into a sequence of contextual representations (\textit{source contents}), and the decoder generates a target sentence word-by-word based on part of the source content assigned by an attention model~\cite{bahdanau2014neural}. 
Like human translators, 
NMT systems should have the ability to know the relevant source-side context for the current word (\present), as well as recognize what parts in the source contents have been translated~(\past) and what parts have not~(\future), at each decoding step. 
Accordingly, the \past, \present\ and \future~are three \textit{dynamically} changing states during the whole translation process.



Previous studies have shown that NMT models are likely to face the illness of inadequate translation~\cite{kong2018neural}, which is usually embodied in over- and under-translation problems~\cite{tu-EtAl:2016:P16-1,tu_etal:17}. This issue may be attributed to the poor ability of NMT of recognizing the dynamic translated and untranslated contents. To remedy this, \newcite{zheng2018modeling} first demonstrate that explicitly tracking \past\ and \future\ contents helps NMT models alleviate this issue and generate better translation. In their work, the running \past\ and \future\ contents are modeled as recurrent states. 
However, the recurrent process is still non-trivial to determine which parts of the source words are the \past\ and which are the \future, and to what extent the recurrent states represent them respectively, this less interpretable nature is probably not the best way to model and exploit the dynamic \past\ and \future.

\begin{figure}[t] 
    \centering
    \includegraphics[width=0.48\textwidth]{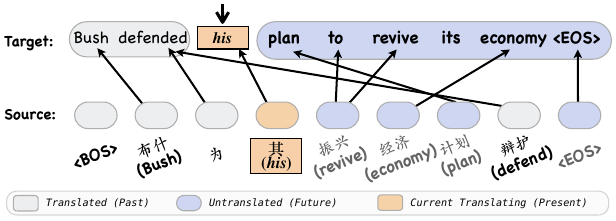}
    \caption{An example of separation of \past~and \future\ in machine translation. When generating the current translation ``his", the source tokens ``$\langle$BOS$\rangle$'', ``布什(Bush)'' and phrase ``为...辩护(defend)'' are the translated contents (\past), while the remaining tokens are untranslated contents (\future). 
    }
    \label{fig:example}
  \end{figure}

We argue that an explicit separation of the source words into two groups, representing \past\ and \future\ respectively  (\Figref{fig:example}), could be more beneficial not only for easy and direct recognition of the translated and untranslated source contents, but also for better interpretation of model's behavior of the recognition.
We formulate the explicit separation as a procedure of parts-to-wholes assignment: the representation of each source words (parts) should be assigned to its associated group of either \past\ or \future\ (wholes).

\begin{CJK}{UTF8}{gbsn}
\end{CJK}

In this paper, we implement this idea using Capsule Network~\cite{hinton2011transforming} with routing-by-agreement mechanism~\cite{sabour2017dynamic}, which has demonstrated its appealing strength of solving the problem of parts-to-wholes assignment~\cite{ hinton2018matrix,gong2018information,dou2019dynamic,li2019information}, to model the separation of the \past\ and \future:
\begin{compactenum}
    \item We first cast the \past\ and \future\ source contents as two groups of capsules. 
    \item We then design a novel variant of the routing-by-agreement mechanism, called {\em Guided Dynamic Routing} ({\sc Gdr}), which is {\em guided} by the current translating status at each decoding step to assign each source word to its associated capsules by assignment probabilities for several routing iterations.
    \item Finally, the \past\ and \future\ capsules accumulate their expected contents from representations, and are fed into the decoder to provide a time-dependent holistic view of context to decide the prediction. 
\end{compactenum}
In addition, two auxiliary learning signals facilitate {\sc Gdr}'s acquiring of our expected functionality, other than implicit learning within the training process of the NMT model. 

We conducted extensive experiments and analysis to verify the effectiveness of our proposed model. Experiments on Chinese-to-English, English-to-German, and English-to-Romanian show consistent and substantial improvements over the Transformer~\cite{Vaswani2017Attention} or \textsc{Rnmt}~\cite{bahdanau2014neural}. Visualized evidence proves that our approach does acquire the expected ability to separate the source words into \past\ and \future, which is highly interpretable. We also observe that our model does alleviate the inadequate translation problem: Human subjective evaluation reveals that our model produces more adequate and high-quality translations than Transformer. Length analysis regarding source sentences shows that our model generates not only longer but also better translations.

\section{Neural Machine Translation}
Neural models for sequence-to-sequence tasks such as machine translation often adopt an \textit{encoder-decoder} framework. Given a source sentence ${\bf x}=\langle x_1, \dots, x_{I}\rangle$, a NMT model learns to predict a target sentence ${\bf y}=\langle y_1, \dots, y_{T}\rangle$ by maximizing the conditional probabilities $p(\mathbf{y}|\mathbf{x}) = \prod_{t=1}^{T} p(y_t|y_{<t},\mathbf{x})$. Specifically, an encoder first maps the source sentence into a sequence of encoded representations:
\begin{equation}
  \mathbf{h} = \langle\vh_1, \dots, \vh_I \rangle = f_{e}(\mathbf{x}),
\end{equation}
where $f_{e}$ is the encoder's transformation function. Given the encoded representations of the source words, a decoder generates the sequence of target words $\mathbf{y}$ autoregressively:
\begin{align}
  \vz_t & = f_{d}(y_{<t}, \va_t) \label{eq:dec-state}, \\ 
  p(y_t|y_{<t}, \mathbf{x}) & = \mathrm{softmax}(E(y_t)^\top \vz_t) \label{eq:gen-probs},
\end{align}
where $E(y_t)$ is the embedding of $y_t$. The current word is predicted based on the decoder state $\vz_t$. $f_{d}$ is the transformation function 
of decoder, which determines $\vz_t$ based on the target translation trajectory $y_{<t}$, and the lexical-level source content $\va_t$ that is most relevant to \present\ translation by an attention model~\cite{bahdanau2014neural}. Ideally, with all the source encoded representations in the encoder, NMT models should be able to update translated and untranslated source contents and keep them in mind. However, most of existing NMT models lack an explicit functionality to maintain the translated and untranslated contents, failing to distinguish  the source words being of either \past\ or \future\ \cite{zheng2018modeling}, which is likely to suffer from severe inadequate translation problem~\cite{tu-EtAl:2016:P16-1,kong2018neural}.



\section{Approach}

\paragraph{Motivation}
Our intuition arises straightforwardly: if we could tell the translated and untranslated source contents apart by directly separating the source words into \past\ and \future~categories at each decoding step, the \present\ translation could benefit from the dynamically holistic context (i.e., \past + \present + \future). To this purpose, we should design a mechanism by which each word could be recognized and assigned to a distinct category, i.e., \past\ or \future\ contents, subject to the translation status at present. This procedure can be seen as a parts-to-wholes assignment, in which the encoder hidden states of the source words (parts) are supposed to be assigned to either \past~or \future~ (wholes). 

Capsule network~\cite{hinton2011transforming} has shown its capability of solving the problem of assigning parts to wholes~\cite{sabour2017dynamic}. A capsule is a vector of neurons which represents different properties of the same entity from the input~\cite{sabour2017dynamic}. The functionality relies on a fast iterative process called routing-by-agreement, whose basic idea is to iteratively refine the proportion of how much a part should be assigned to a whole, based on the agreement between the part and the whole~\cite{dou2019dynamic}. Therefore, it is appealing to investigate if this mechanism could be employed for our intuition.

\subsection{Guided Dynamic Routing (\textsc{Gdr})} 
\begin{figure}[t] 
  \centering
  \includegraphics[width=0.49\textwidth]{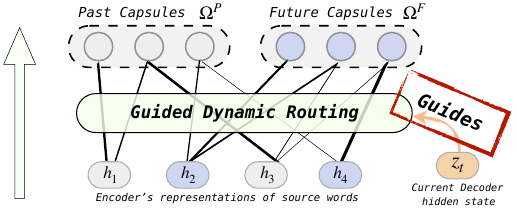}
  \caption{Illustration of the Guided Dynamic Routing.}
  \label{fig:gdr}
\end{figure}

Dynamic routing~\cite{sabour2017dynamic} is an implementation of routing-by-agreement, where it runs intrinsically without any external guidance. However, what we expect is a mechanism driven by the decoding status at present. Here we propose a variant of dynamic routing mechanism called {\it Guided Dynamic Routing} ({\sc Gdr}), where the routing process is \textit{guided} by the translating information at each decoding step (Figure \ref{fig:gdr}).

Formally, we cast the source encoded representations $\rvh$ of $I$ source words to be input capsules, while we denote $\mathbf{\Omega}$ as output capsules, which consist of $J$ entries. Initially, we assume that $J/2$ of them ($\mathbf{\Omega}^P$) represent the \past~contents, and the rest $J/2$ capsules ($\mathbf{\Omega}^F$) represent the \future:
\begin{align}
    \mathbf{\Omega}^P = \langle \Omega^P_1, \cdots, \Omega^P_{J/2} \rangle, ~~~ \mathbf{\Omega}^F = \langle \Omega^F_1, \cdots, \Omega^F_{J/2} \rangle \nonumber .
\end{align}
where each capsule is represented by a $d_c$-dimension vector. We assemble these \past\ and \future\ capsules together, which are expected to competing for source information, i.e., we now have $\mathbf{\Omega} = \mathbf{\Omega}^P \cup \mathbf{\Omega}^F$. We will describe how to teach these capsules to retrieve their relevant parts from source contents in the Section \ref{sec:aux}.  
{\textit{\textbf{Note}}} that we employ \textsc{Gdr} at every decoding step $t$ to obtain the time-dependent \past\ and \future\, and omit the subscript $t$ for simplicity.

In the dynamic routing process, each vector output of capsule $j$ is calculated with a non-linear {\it squashing} function~\cite{sabour2017dynamic}:
\begin{align}
  \Omega_j &= \frac{\| \vs_j \|^2}{1+\| \vs_j\|^2} \frac{ \vs_j}{\| \vs_j \|},  ~~~ \vs_j &= \sum_i^I c_{ij} \vv_{ij} \label{eq:dr-out1},
\end{align}
where $\vs_j$ is the accumulated input of capsule $\Omega_j$, which is a weighted sum over all {\it vote vectors} $\vv_{ij}$. $\vv_{ij}$ is transformed from the input capsule $\vh_i$:
\begin{align}
  \vv_{ij} = \mW_j \vh_i,
\end{align}
where $\mW_j \in \mathbb{R}^{d \times d_c}$ is a trainable matrix for $j$-th output capsule\footnote{Note that unlike \newcite{sabour2017dynamic}, where each pair of input capsule $i$ and output capsule $j$ has a distinct transformation matrix $\mW_{ij}$ as their numbers are predefined ($I\times J$ transformation matrices in total), here we share the transformation matrix $\mW_j$ of output capsule $j$ among all the input capsules due to the varied amount of the source words. So there are $J$ transformation matrices in our model.}. 
$c_{ij}$ is the assignment probability (i.e. the agreement) that is determined by the iterative dynamic routing.
The assignment probabilities $c_{i\cdot}$ associated with each input capsule $\vh_i$ sum to 1: $\sum_j c_{ij} = 1$, and are computed by:
\begin{equation}
  c_{ij} = \mathrm{softmax}(b_{ij}) , \label{eq:rouing-sm}
\end{equation}  
where routing logit $b_{ij}$ is initialized as all 0s, which measures the degree that $\vh_i$ should be sent to $\Omega_j$. The initial assignment probabilities are then iteratively updated by measuring the agreement between the vote vector $\vv_{ij} $ and capsule $\Omega_j$ by an MLP, considering the current decoding state $\vz_t$: 
\begin{equation}
  b_{ij} \leftarrow b_{ij} + \vw^{\top}\mathrm{tanh}(\mW_b [\vz_t; \vv_{ij}; \Omega_j]) , \label{eq:dr-b}
\end{equation}  
where $\mW_b \in \mathbb{R}^{d+d_c*2}$ and $\vw \in \mathbb{R}^{d_c}$ are learnable parameters. Instead of using simple scalar product, i.e., $b_{ij} = \vv_{ij}^{\top} \Omega_j$~\cite{sabour2017dynamic}, which could not consider the current decoding state as a condition signal, we resort to the MLP to take $\vz_i$ into account inspired by MLP-based attention mechanism~\cite{bahdanau2014neural,D15-1166}. That is why we call it ``guided'' dynamic routing.

Now with the awareness of the current decoding status, the hidden state (input capsule) of a source word prefers to send its representation to the output capsules, which have large routing agreements associated with the input capsule. After a few rounds of iterations, the output capsules are able to ignore all but the most relevant information from the source hidden states, representing a distinct aspect of either \past\ or \future.

\paragraph{Redundant Capsules}
In some cases, some parts of the source sentence may belong to neither past contents nor future contents. For example, function words in English (e.g., ``the'') could not find its counterpart translation in Chinese. 
Therefore, we add additional Redundant Capsules $\mathbf{\Omega}^R$ (also known as ``orphan capsules'' in \newcite{sabour2017dynamic}), which are expected to receive higher routing assignment probabilities when a source word should not belong to either \past~or \future.

      
   

We show the algorithm of our guided dynamic routing in Algorithm~\ref{alg:dynamic-routing}.

\begin{algorithm}[tb]
    \footnotesize
  \caption{Guided Dynamic Routing ({\sc Gdr})}
  \label{alg:dynamic-routing}
  \hspace*{0.02in} {\bf Input:} Encoder hidden state $\textbf{h}$, current decoding hidden state  $\vz_t$, and number of routing iterations $r$. \\
  \hspace*{0.02in} {\bf Output:} \past, \future, and redundant capsules. \\
  \hspace*{0.02in} {\bf procedure:} \textsc{Gdr}($\textbf{h}$, $\vz_t$, $r$)
  \begin{algorithmic}[1] 
\State $\forall i \in \rvh, j \in \mathbf{\Omega}: b_{ij} \leftarrow 0, \vv_{ij} \leftarrow \mW_j \vh_i$ \Comment{\textit{Initializing routing logits, and vote vectors.}}
  \For{$r$ iterations} 
      \State $\forall i \in \rvh, j \in \mathbf{\Omega}$: Compute assign. probs. $c_{ij}$ by Eq.~\ref{eq:rouing-sm} 
      \State $\forall j \in \mathbf{\Omega}:$ Compute capsules $\Omega_j$ by Eq.~\ref{eq:dr-out1}
      \State $\forall i \in \rvh, j \in \mathbf{\Omega}:$ Update routing logits $b_{ij}$ by Eq.~\ref{eq:dr-b}
  \EndFor 
   
  \State $[\mathbf{\Omega}^P; \mathbf{\Omega}^F; \mathbf{\Omega}^R] = \mathbf{\Omega}$  \Comment{\textit{Return past, future, and redundant capsules}}
  \State \textbf{return} $\mathbf{\Omega}^P, \mathbf{\Omega}^F, \mathbf{\Omega}^R$ 
  \end{algorithmic}
\end{algorithm}

\subsection{Integrating into NMT}

\begin{figure}[t] 
  \centering
  \includegraphics[width=0.49\textwidth]{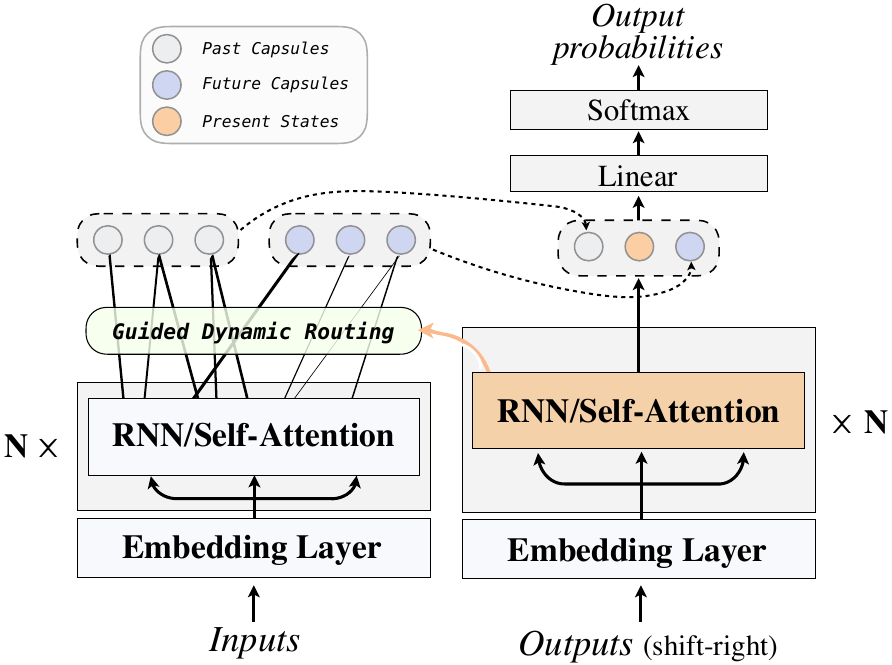}
  \caption{Illustration of our architecture.}
  \label{fig:model}
\end{figure}

The proposed \textsc{Gdr} can be applied on the top of any sequence-to-sequence architecture, which does not require any specific modification. Let us take a Transformer-fashion architecture as example (Figure~\ref{fig:model}).
Given a sentence ${\bf x}=\langle x_1, \dots, x_{I}\rangle$, the encoder leverages $N$ stacked identical layers to map the sentence into contextual representations:
\begin{align}
    \mathbf{h}^{(l)} = \mathrm{EncoderLayer}(\mathbf{{\rvh}}^{(l-1)})  \nonumber,
\end{align}
where the superscript $l$ indicates layer depth.
Based on the encoded source representations $\mathbf{h}^{N}$, a decoder generates translation word by word. The decoder also has $N$ stacked identical layers:
\begin{align}
  \mathbf{z}^{(l)} = \mathrm{DecoderLayer}(\mathbf{z}^{(l-1)}, \mathbf{a}^{(l)})  \nonumber, \\
  \mathbf{a}^{(l)} = \mathrm{Attention}(\mathbf{z}^{(l-1)}, \mathbf{h}^{(N)}) \nonumber ,
\end{align} 
where $\mathbf{a}^{(l)}$ is the lexical-level source context assigned by an attention mechanism between current decoder layer and the last encoder layer. Given the hidden states of the last decoder layer $\mathbf{z}^{(N)}$, we perform our proposed guided dynamic routing ({\sc Gdr}) mechanism to compute the \past~and \future~contents from the source side and obtain the holistic context of each decoding step:
\begin{align}
  &\mathbf{\Omega}^P, \mathbf{\Omega}^F, \mathbf{\Omega}^R = \textsc{Gdr}(\mathbf{z}^{(N)}, \mathbf{h}^{(N)})  \nonumber, \\
  &\mathbf{o} = \mathrm{FeedForward}(\mathbf{z}^{(N)}, \mathbf{\Omega}^P, \mathbf{\Omega}^F) + \mathbf{z}^{(N)} \nonumber,
\end{align} 
where $\mathbf{o} = \langle \vo_1, \cdots, \vo_T \rangle$ is the sequence of the holistic context of each decoding step. Based on the holistic context, the output probabilities are computed as:
\begin{align}
  p(y_t|y_{\le t}, \mathbf{x}) = \mathrm{softmax}(g(\vo_t)). \nonumber
\end{align}
The NMT model is now able to employ the dynamic holistic context for better generation.

\subsection{Learning \past\ and \future\ as Expected}
\subsubsection*{Auxiliary Guided Losses}
\label{sec:aux}
To ensure that the dynamic routing process runs as expected, we introduce the following auxiliary guided signals to assist the learning process.

\paragraph{Bag-of-Word Constraint} \newcite{Weng:2017:EMNLP} propose a multitasking scheme to boost NMT by predicting the bag-of-words of target sentence using the Word Predictions approach. 
Inspired by this work, we introduce a \textsc{BoW} constraint to encourage the \past\ and \future\ capsules to be predictive of the preceding and subsequent bag-of-words regarding each decoding step respectively:
\begin{align}
  \mathcal{L}_\textsc{BoW} & = \frac{1}{T} \sum_{t=0}^T \big( - \log p_\textsc{pre}(y_{\le t}|\mathbf{\Omega}^P_t) \nonumber \\ 
  &~~~~~~~~~~~~~~~ - \log p_\textsc{sub}(y_{\ge t}|\mathbf{\Omega}^F_t) \big) \nonumber,
\end{align}
where $p_{\sc pre}(y_{\le t}|\mathbf{\Omega}^P_t)$ and $p_{\sc sub}(y_{\ge t}|\mathbf{\Omega}^F_t)$ are the predicted probabilities of the preceding bag-of-words and subsequent words at decoding step $t$, respectively. For instance, the probabilities of the preceding bag-of-words are computed by:
\begin{align}
  p_\textsc{pre}(y_{<t}|\mathbf{\Omega}^P_t) &= \prod_{\tau \in [1, t]} p_\textsc{pre}(y_{\tau}|\mathbf{\Omega}^P_t) \nonumber\\[-2pt]
  &\propto \prod_{\tau \in [1, t]} \exp (\mE(y_\tau)^\top \mW_{\textsc{Bow}}^P \mathbf{\Omega}^P_t) \nonumber.
\end{align} 
The computation of $p_\textsc{sub}(y_{\ge t}|\mathbf{\Omega}^F_t)$ is similar. By applying the \textsc{BoW} constraint, the \textsc{Past} and \textsc{Future} capsules can learn to reflect the target-side past and future bag-of-words information.

\paragraph{Bilingual Content Agreement}
Intuitively, the translated source contents should be semantically equivalent to the translated target contents, and so do untranslated contents. Thus, a natural idea is to encourage the source \textsc{Past} contents, modeled by the \textsc{Past} capsule to be close to the target \past\ representation at each decoding step, and the same for the \textsc{Future}. Hence, we introduce a Bilingual Content Agreement (\textsc{Bca}) to require the bilingual semantic-equivalent contents to be predictive to each other by Minimum Square Estimation (MSE) loss:
\begin{align}
  \mathcal{L}_\textsc{Bca} \!&= \!\frac{1}{T} \sum_{t=1}^{T} \|\mathbf{\Omega}^P_t \!-\! \mW^{P}_{\textsc{Bca}}(\frac{1}{t} \sum_{\tau=1}^{t} \vz_\tau) \|^2  \nonumber \\[-6pt]
  & ~~~~~~~~~~~~~+ \|\mathbf{\Omega}^F_t \!-\! \mW^F_{\textsc{Bca}}(\frac{1}{T\!-\!t\!+\!1} \sum_{\tau=t}^{T} \vz_\tau)\|^2,  \nonumber
\end{align}
where the target-side past information is represented by the averaged results of the decoder hidden states of all preceding words, while the average of subsequent decoder hidden states represents the target-side future information.

\subsubsection*{Training}
Given the dataset of parallel training examples $\{\langle \mathbf{x}^{(m)}, \mathbf{y}^{(m)}\rangle \}_{m=1}^{M}$, the model parameters are trained by minimizing the loss $\mathcal{L}(\vtheta)$, where $\vtheta$ is the set of all the parameter of the proposed model:
\begin{align}
\label{eq:new-obj}
  \mathcal{L}(\vtheta)\!=\!\frac{1}{M}\!\!\sum_{m=1}^{M}\!\big(&-\!\log p(\mathbf{y}^{(m)}|\mathbf{x}^{(m)}) \nonumber \\[-6pt]
  &+\lambda_1 \cdot \mathcal{L}_{\rm BoW} +\lambda_2 \cdot \mathcal{L}_{\textsc{Bca}} \big) \nonumber,
\end{align}
where $\lambda_1$ and $\lambda_2$ are hyper-parameters. 




\section{Experiment}
We mainly evaluated our approaches on the widely used NIST Chinese-to-English (Zh-En) translation task. We also conducted translation experiments on WMT14 English-to-German~(En-De) and WMT16 English-to-Romanian~(En-Ro):

{1. NIST Zh-En.}
The training data consists of 1.09 million sentence pairs extracted from LDC\footnote{The corpora includes LDC2002E18, LDC2003E07, LDC2003E14, Hansards portion of LDC2004T07, LDC2004T08 and LDC2005T06}. We used NIST {MT03} as the development set (Dev); MT04, {MT05}, {MT06} as the test sets.  

{2. WMT14 En-De.}
The training data consists of 4.5 million sentence pairs from WMT14 news translation task. We used newstest2013 as the development set and newstest2014 as the test set.

{3. WMT16 En-Ro.}
The training data consists of 0.6 million sentence pairs from WMT16 news translation task. We used newstest2015 as the development set and newstest2016 as the test set. 

We used \texttt{transformer{\_}base} configuration~\cite{Vaswani2017Attention} for all the models. We run the dynamic routing for $r\!\!=\!\!3$ iterations. The dimension $d_c$ of a single capsule is 256. Either \past\ or \future\ content was represented by $\frac{J}{2}=2$ capsules. Our proposed models were trained on the top of pre-trained baseline models\footnote{Pre-training is only for efficiency purpose. Our approach could also learn from scratch.}. $\lambda_1$ and $\lambda_2$ in training objective were set to 1.
In Appendix, we provide details for the training settings.

\begin{table*}[t] 
  \centering
  \resizebox{\textwidth}{!}{%
  \begin{tabular}{lccc|ccccc} 
  \toprule
  {\bf Model}                                                               & $|\vtheta|$& $v_{\rm{train}}$ & $v_{\rm{test}}$    & Dev   & MT04  & MT05  & MT06  & Tests Avg. \\ 
    \hline
  Transformer                                                               & 66.1m & 1.00$\times$  & 1.00$\times$  & 45.83 & 46.66 & 43.36 & 42.17 & 44.06     \\
  \hline
  \textsc{Gdr}                                                              & 68.9m & 0.77$\times$  & 0.94$\times$  & 46.50 & 47.03 & 45.50 & 42.21 & 44.91 (+0.75)  \\
  ~~~~ + $\mathcal{L}_\textsc{BoW}$                                           & 69.2m & 0.70$\times$  & 0.94$\times$  & 47.12 & 48.09 & 45.98 & 42.68 & 45.58 (+1.42) \\
  ~~~~ + $\mathcal{L}_\textsc{Bca}$                                           & 69.4m & 0.75$\times$  & 0.94$\times$  & 46.86 & 48.00 & 45.67 & 42.62 & 45.43 (+1.37) \\
  ~~~~ + $\mathcal{L}_\textsc{BoW}$ + $\mathcal{L}_\textsc{Bca}$ [{\sc Ours}]   & 69.7m & 0.67$\times$  & 0.94$\times$  & \bf 47.52 & \bf 48.13 & \bf 45.98 & \bf 42.85 & \bf 45.65 (+1.59)   \\
  \textsc{Ours} - {\it redundant capsules}                                           & 68.7m & 0.69$\times$  & 0.94$\times$  & 47.20 & 47.82 & 45.59 & 42.51 & 45.30 (+1.24) \\
  \hline
  \hline
  \textsc{Rnmt}                                                                      & 50.2m & 1.00$\times$  & 1.00$\times$  & 35.98 & 37.85 & 36.12 & 35.86 & 36.61    \\
  ~~~~ +\textsc{PFRnn} \cite{zheng2018modeling}                                   & N/A   & 0.54$\times$  & 0.74$\times$  & 37.90 & 40.37 & 36.75 & 36.44 &  37.85  (+1.24)            \\
  ~~~~ +\textsc{Aol} \cite{kong2018neural}                                        & N/A   & 0.57$\times$  & 1.00$\times$          & 37.61 & 40.05 & 37.58 & 36.87 & 38.16    (+1.55)           \\
  \hline
  {\sc Ours}  & 53.9m & 0.62$\times$  & 0.90$\times$  & \bf 38.10 & \bf 40.87 & \bf 37.50 & \bf 37.00 & \bf 38.45 (+1.84) \\
  \hline
  \end{tabular} 
  }
  \caption{Experiment ressuts on NIST Zh-En task, including number of parameters ($|\vtheta|$, excluding word embeddings), training/testing speeds ($v_{\rm{train}}$/$v_{\rm{test}}$), and translation results in case-insensitive BLEU. }
  \label{tab:zh-en_bleu} 
\end{table*}

\subsection{NIST Zh-En Translation}
We list the results of our experiments on NIST Zh-En task in Table~\ref{tab:zh-en_bleu} concerning two different architectures, i.e., Transformer and \textsc{Rnmt}. As we can see, all of our models substantially outperform the baselines in terms of averaged BLEU score of all the test sets. Among them, our best model achieves 45.65 BLEU based on Transformer architecture. We also find that redundant capsules are helpful while discarding them leads to -0.35 BLEU degradation (45.65 vs 45.30).

\paragraph{Architectures} Our approach shows consistent effects on both Transformer and RNMT architectures. In comparison to the Transformer baseline, our model achieves at most +1.59 BLEU improvement (45.65 v.s 44.06), while +1.84 BLEU improvement over RNMT baselines (38.45 v.s 36.61). These results indicate the compatibility of our approach to different architectures. 

\paragraph{Auxiliary Guided Losses} 
Both the auxiliary guided losses help our model for better learning. The \textsc{BoW} constraint leads to a +0.67 improvement compared to the vanilla \textsc{Gdr}, while the benefit is +0.62 for \textsc{Bca}. Combination of both gains the most margins (+0.84), which means that they can supplement each other.

\paragraph{Efficiency} To examine the efficiency of the proposed approach, we also list the relative speed of both training and testing. Our approach is 0.67$\times$ slower than the Transformer baseline in training phase, however, it does not hurt the speed of testing too much (0.94$\times$). It is because the most extra computation in training phrase is related to the softmax operations of \textsc{BoW} losses, the degradation of the testing efficiency is moderate.

\paragraph{Comparison to Other Work} On the experiments on \textsc{Rnmt} architecture, we list two related works. \newcite{zheng2018modeling} use extra \past\ and \future\ RNNs to capture translated and untranslated contents recurrently (\textsc{PFRnn}), while \newcite{kong2018neural} directly leverage translation adequacy as learning reward by their proposed Adequacy-oriented Learning (\textsc{Aol}). Compared to them, our model also enjoys competitive improvements due to the explicit separation of source contents. In addition, \textsc{PFRnn} is non-trivial to adapt to Transformer, because it requires a recurrent process which fails to be compatible with parallel training of Transformer, scarifying Transformer's efficiency advantage.




\begin{table}[t]
  \footnotesize
     \centering
  \resizebox{0.49\textwidth}{!}{%
  \begin{tabular}{lcc}
  \toprule
      {\bf Model}                                           & En-De & En-Ro    \\
      
      \hline
      GNMT+RL \cite{wu2016google}                           & 24.6  & N/A\\
      ConvS2S \cite{Gehring2017ConSeq}                      & 25.2  & 29.88 \\
      Transformer \cite{Vaswani2017Attention}               & 27.3  & N/A\\
      ~~~~ +\textsc{Aol}~\cite{kong2018neural}             & 28.01 & N/A \\
      Transformer~\cite{Gu2017NAT}                          & N/A   & 31.91 \\
      \hline
      Transformer                                           & 27.14 & 32.10 \\
      {\sc Ours}                                            & \bf 28.10 & \bf 32.96 \\
      \hline
  \end{tabular}
  }
  \caption{Case-sensitive BLEU on WMT14 En-De and WMT16 En-Ro tasks.}
  \label{tab:wmt}
\end{table}

\subsection{WMT En-De and En-Ro Translation}

We evaluated our approach on WMT14 En-De and WMT16 En-Ro tasks. As shown in Table \ref{tab:wmt}, our reproduced Transformer baseline systems are close to the state-of-the-art results in previous work, which guarantee the comparability of our experiments. The results show a consistent trend of improvements as NIST Zh-En task on WMT14 En-De (+0.96 BLEU) and WMT16 En-Ro (+0.86 BLEU) benchmarks. We also list the results of other published research for comparison, where our model outperforms the previous results in both language pairs. Note that our approach also surpasses \newcite{kong2018neural} on WMT14 En-De task. These experiments demonstrate the effectiveness of our approach across different language pairs.

\subsection{Analysis and Discussion}


\begin{CJK}{UTF8}{gbsn}
\begin{figure*}[t] 
  \centering
  \includegraphics[width=0.785\textwidth]{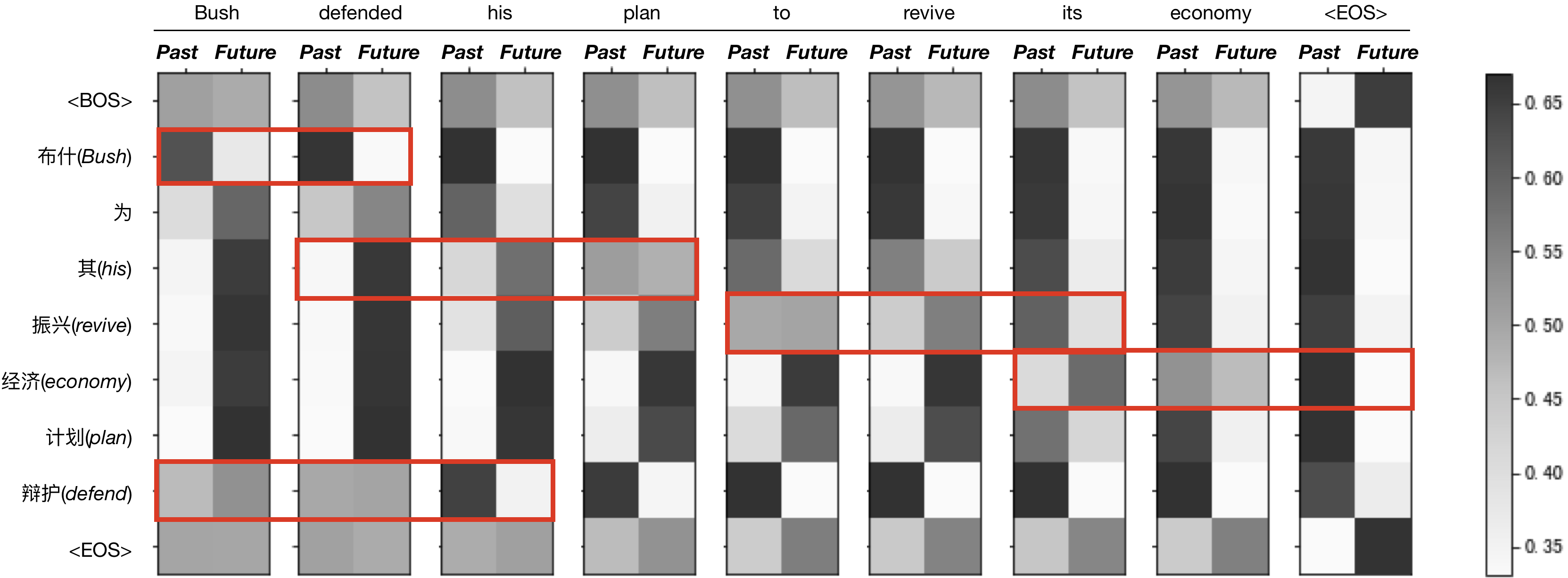}
  \caption{Visualization of the assignment probabilities of iterative routing. Each sub-heatmap is associated with a target word, where the left column is the probabilities of each source words routing to the \past\ capsules, and the right one is to the \future\. Examples in the red frame indicate the changes before and after the generation of the central word. We omit the assignment probabilities associated with the redundant capsules for simplicity. For instance, after the target word ``defended'' was generated, the assignment probabilities of its source translation ``辩护'' changed from \future\ to \past. Results of ``Bush``, ``his'', ``revive'' and ``economy'' are similar, except a adverse case (``plan'').}
  \label{fig:case-visualization}
\end{figure*}
\end{CJK}

\paragraph{\textit{\textbf{Our model learns \past~and \future.}}}
\begin{CJK}{UTF8}{gbsn}
We visualize the assignment probabilities in the last routing iteration (\Figref{fig:case-visualization}). Interestingly, there is a clear trend that the assignment probabilities to the \past\ capsules gradually raise up, while those to the \future\ capsules reduce to around zeros. This phenomenon is consistent with the intuition that the translated contents should aggregate and the untranslated should decline~\cite{zheng2018modeling}. The assignment weights of a specific word change from \future\ to \past\ after being generated.
These pieces of evidence give a strong verification that our \textsc{Gdr} mechanism indeed has learned to distinguish the \past\ contents and \future\ contents in the source-side. 
\end{CJK}


Moreover, we measure how well our capsules accumulate the expected contents by comparison between the \textsc{BoW} predictions and ground-truth target words. Accordingly, we define a \textit{top-5 overlap rate} ($r_{\textsc{ol}}$) for predicting preceding and subsequent words are defined as follow, respectively:  
   $r_{\textsc{ol}}^P\!\!\! = \frac{1}{T} \sum_{t=1}^T \frac{|\mathrm{Top_{5t}}(p_{\sc pre}(\mathbf{\Omega}_t^P)) \cap y_{<=t}|}{|y_{<=t}|},~~
   r_{\textsc{ol}}^F\!\!\! = \frac{1}{T} \sum_{t=1}^T \frac{|\mathrm{Top_{5(T-t)}}(p_{\sc sub}(\mathbf{\Omega}_t^F)) \cap y_{>=t}|}{|y_{>=t}|}.$
The \past\ capsules achieves $r_{\textsc{ol}}^P$ of 0.72, while $r_{\textsc{ol}}^F$ of 0.70 for the \future\ capsules. The results indicate that the capsules could predict the corresponding words to a certain extent, which implies the capsules contain the expected information of \past\ or \future\ contents.

\begin{table}[t]
\centering
\footnotesize
\begin{tabular}{lcc}
\toprule
  \bf Model                 & ~~~~~{Transformer}~~~~~ & ~~~~~\textsc{Ours}~~~~~      \\
\hline
\textsc{Cdr}       & 0.73                 & \bf 0.79           \\
\hline
\hline
\multicolumn{3}{c}{\textsc{Human Evaluation}}                  \\
\hline
\textsc{Quality}   & 4.39$\pm$.11         & \bf 4.66$\pm$.10   \\
\textsc{Over(\%)}  & 0.03$\pm$.01         & \bf 0.01$\pm$.01   \\
\textsc{Under(\%)} & 3.83$\pm$.97         & \bf 2.41$\pm$.80  \\
\hline
\end{tabular}%
\caption{Evaluation on translation quality and adequacy. For {\sc Human} evaluation, we asked three evaluators to score translations from 100 source sentences, which are randomly sampled from the testsets from anonymous systems, the {\sc Quality} from 1 to 5 (higher is better), and the proportions of source words concerning {\sc Over}- and {\sc Under}-translation, respectively. }
\label{tab:adequacy}
\end{table}

\paragraph{\textit{\textbf{Translations become better and more adequate.}}}
To validate the translation adequacy of our model, we use Coverage Difference Ratio ({\sc Cdr})  proposed by \newcite{kong2018neural}, i.e., $\textsc{Cdr} = 1 - \frac{|C_{\rm ref} \setminus C_{\rm gen}|}{| C_{\rm ref}|} $, where $C_{\rm ref}$ and $C_{\rm gen}$ are the set of source words covered by the reference and translation, respectively. The {\sc Cdr} reflects the translation adequacy by comparing the source coverages between reference and translation. As shown in Table \ref{tab:adequacy}, our approach achieves a better \textsc{Cdr} than the Transformer baseline, which means superiority in translation adequacy.

Following~\newcite{zheng2018modeling}, we also conduct subjective evaluations to validate the benefit of modeling \textsc{Past} and \textsc{Future}~(the last three rows of Table \ref{tab:adequacy}). Surprisingly, we find that the modern NMT model, i.e., Transformer, rarely produces over-translation but still suffers from under-translation. Our model obtains the highest human rating on translation quality while substantially alleviates the under-translation problem than Transformer.


\begin{figure}[t]
  \centering
  \subfloat[Translation length v.s source length]{
  \label{fig:length-length}
  \includegraphics[width=0.4\textwidth]{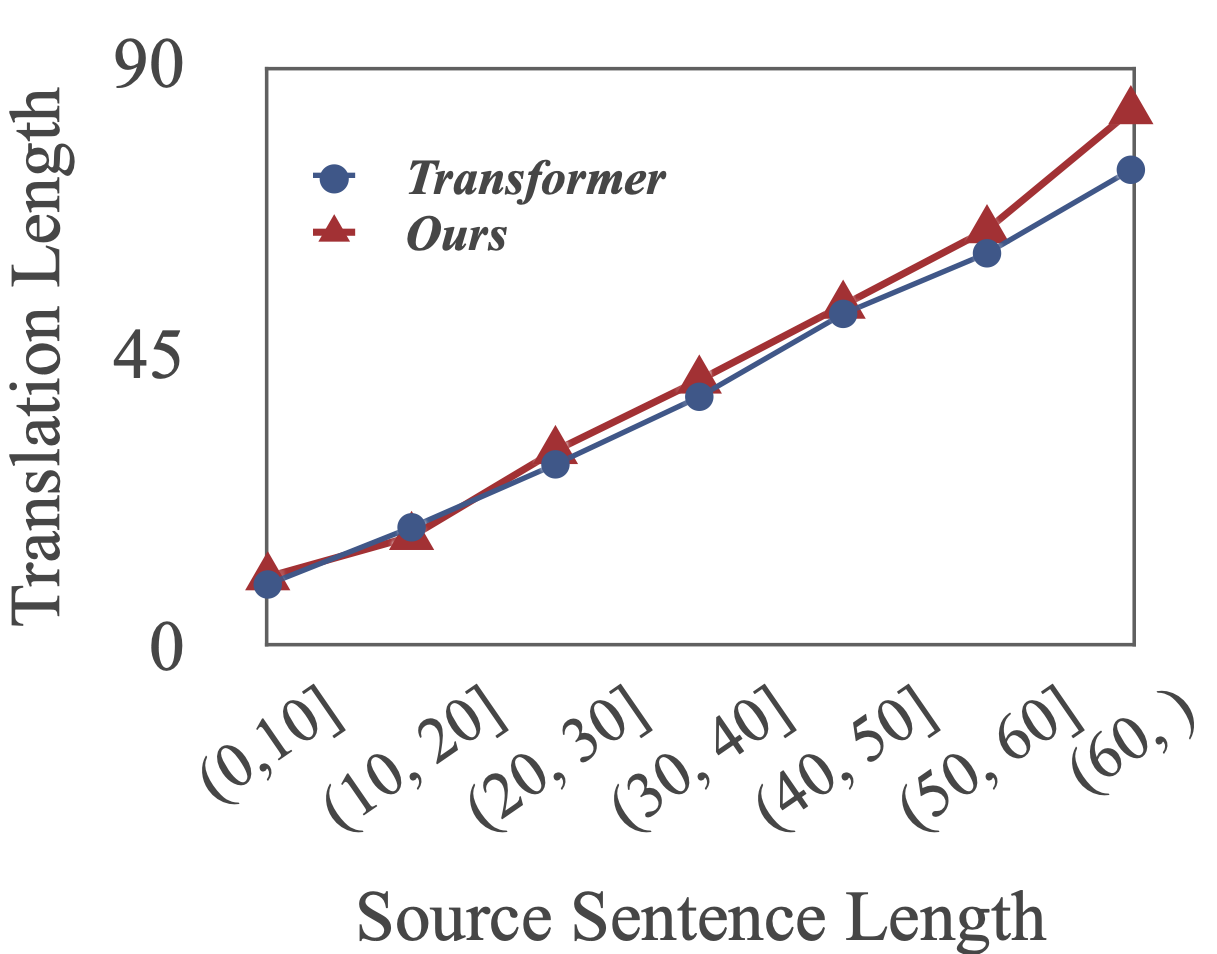}} \\
  \subfloat[BLEU v.s source length]{
  \label{fig:bleu-length}
  \includegraphics[width=0.4\textwidth]{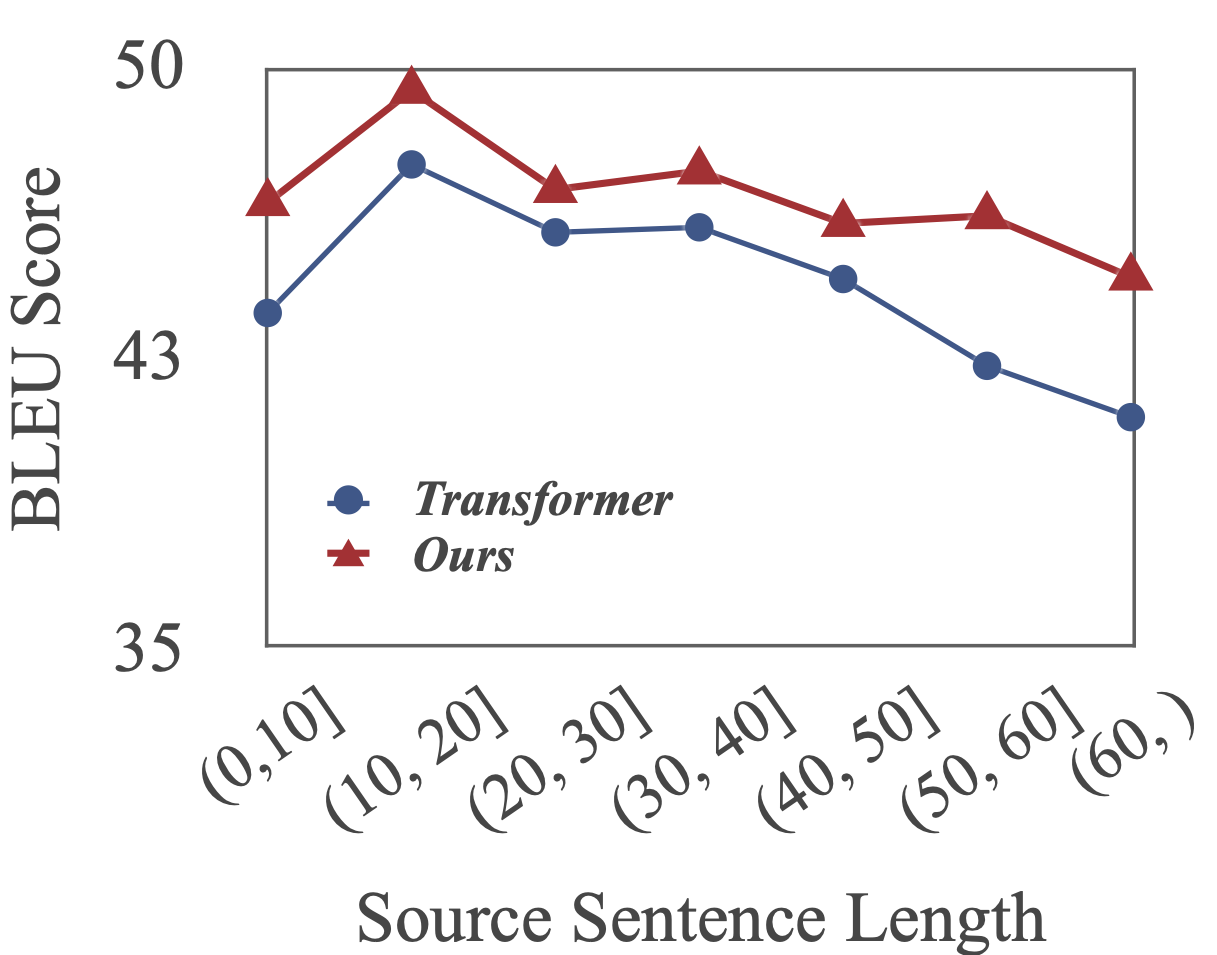}}
  \caption{Comparison regarding source length.}
  \label{fig:length}
  \end{figure}

\paragraph{\textit{\textbf{Longer sentences benefit much more.}}}

We report the comparison with sentence lengths (Figure~\ref{fig:length}). In all the intervals of length, our model does generate better~(Figure \ref{fig:bleu-length}) and longer~(Figure \ref{fig:length-length}) translations.  Interestingly, our approach gets a larger improvement when the input sentences become longer, which are commonly thought hard to translate. We attribute this to the less number of under-translation cases in our model, meaning that our model learns better translation quality and adequacy, especially for long sentences.

\paragraph{\textit{\textbf{Does guided dynamic routing really matter?}}}


Despite the promising numbers of the \textsc{Gdr} and the auxiliary guided losses, a straightforward question rises: will other more simple models also work if they are just equipped with the guided losses to recognize \past\ and \future\ contents? In other word, does the proposed guided dynamic routing really matter?

\begin{figure}[t] 
  ~~\includegraphics[width=0.44\textwidth]{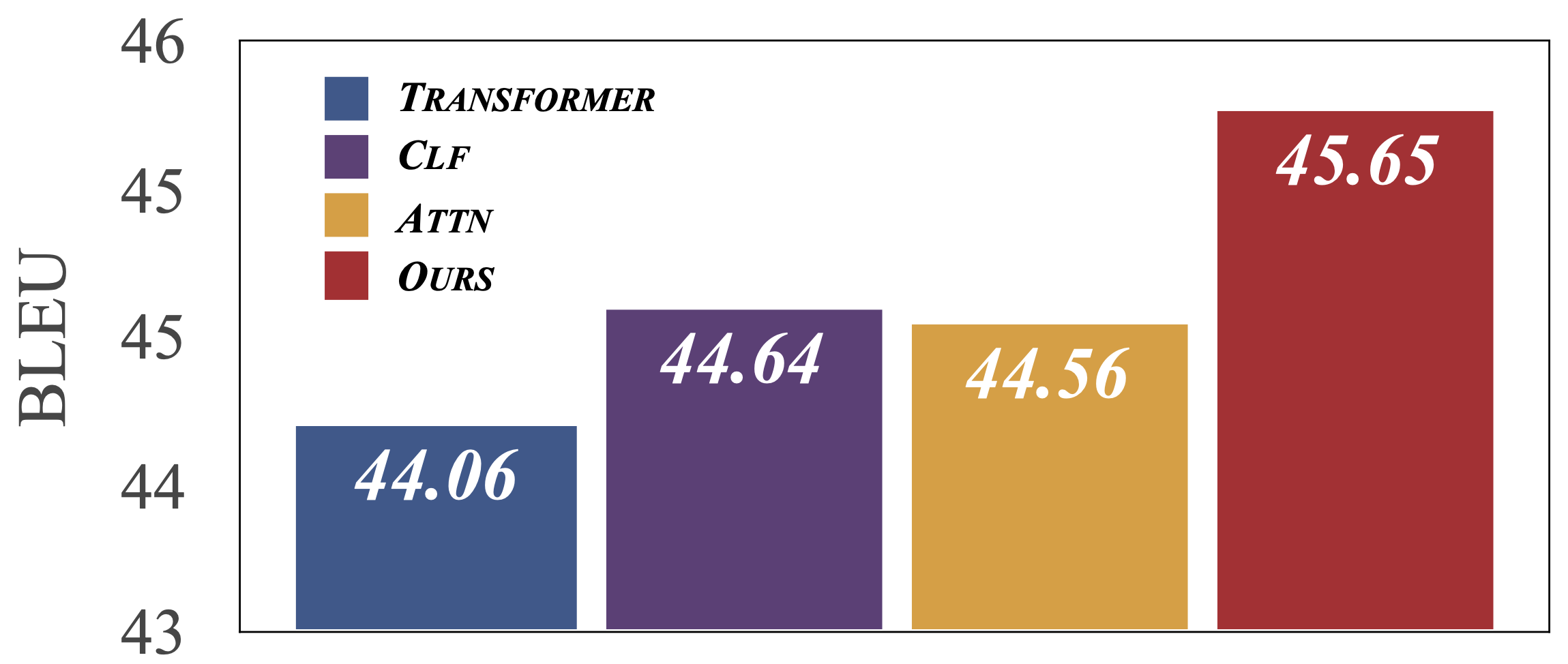}
 \caption{Comparison with simple baselines with the same auxiliary guided loss on NIST Zh-En.}
  \label{fig:guided-comparison}
\end{figure}

To answer this question, we integrate the proposed auxiliary losses into two simple baselines to guide the recognition of past and future: A MLP classifier model (\textsc{Clf}) that determines if a source word is a past word, otherwise future\footnote{\textsc{Clf} is a 3-way classifier that computes the probabilities $p^P(x_i)$, $p^F(x_i)$ and $p^R(x_i)$ (they sum to 1) as past, future and redundant weights, which is similar to Equation \ref{eq:rouing-sm}. The \past\ and \future\ representations are computed by weighted summation, which is similar to Equation \ref{eq:dr-out1}.}; and an attention-based model (\textsc{Attn}) that uses two individual attention modules to retrieve past or future parts from the source words. As shown in Table \ref{fig:guided-comparison}, surprisingly, the simple baselines can obtain improvements, emphasizing the function of the proposed guided losses, while there remain a considerable gaps between our model and them. 
In fact, the \textsc{Clf} is essentially a \textit{one-iteration} variant of \textsc{Gdr}, and iterative refinement by multiple iterations is necessary and effective\footnote{See Appendix for analysis of iteration numbers.}. And the attention mechanism is used for feature pooling, not suitable for parts-to-wholes assignment\footnote{Consider an extreme case that in the end of translation, there is no \future\ content left, but the attention model still produces a \textit{weighted average} over all the source representations, which is nonsense. In contrast, the \textsc{Gdr} is able to assign zero probabilities to the \future\ capsules, solving the source of the problem.  }. 
These experiments reveal that our guided dynamic routing is a better choice to model and exploit the dynamic \past\ and \future.




\section{Related Work}
Inadequate translation problem is a widely known weakness of NMT models, especially when translating long sentences~\cite{kong2018neural,tu-EtAl:2016:P16-1,revisit2019lei}. To alleviate this problem, one direction is to recognize the translated and untranslated contents, and pay more attention to untranslated parts. \newcite{tu-EtAl:2016:P16-1}, \newcite{Mi2016} and \newcite{li2018simple} employ coverage vector or coverage ratio to indicate the lexical-level coverage of source words. \newcite{meng2018neural} influence the attentive vectors by translated/untranslated information. Our work mainly follows the path of \newcite{zheng2018modeling}, which introduce two extra recurrent layers in the decoder to maintain the representations of the past and future translation contents. However, it may be not easy to show the direct correspondence between the source contents and learned representations in the past/future RNN layers, nor compatible with the state-of-the-art Transformer for the additional recurrences prevent Transformer decoder from being parallelized. 

Another direction is to introduce global representations. \newcite{lin2018deconvolution} model a global source representation by deconvolution networks. \newcite{xia2017deliberation,zhang2018asynchronous,geng2018adaptive} propose to provide a holistic view of target sentence by multi-pass decoding. \newcite{zhou2019synchronous} improve \newcite{zhang2018asynchronous} to a synchronous bidirectional decoding fashion. Similarly, \newcite{weng2019correct} deploy bidirectional decoding in interactive translation setting. Different from these work aiming at providing static global information in the whole translation process, our approach models a dynamically global (holistic) context by using capsules network to separate source contents at every decoding steps.


Other efforts explore exploiting future hints. \newcite{serdyuk2018twin} design a Twin Regularization to encourage the hidden states in forward decoder RNN to estimate the representations of a backward RNN. \newcite{Weng:2017:EMNLP} require the decoder states to not only generate the current word but also predict the remain untranslated words. Actor-critic algorithms are employed to predict future properties \cite{Li:2017:arXiv, Bahdanau:2017:ICLR, he2017decoding} by estimating the future rewards for decision making. \newcite{kong2018neural} propose a policy gradient based adequacy-oriented approach to improve translation adequacy. These methods use future information only at the training stage, while our model could also exploit past and future information at inference, which provides accessible clues of translated and untranslated contents. 

Capsule networks~\cite{hinton2011transforming} and its associated assignment policy of dynamic routing~\cite{hinton2011transforming} and EM-routing~\cite{hinton2018matrix} aims at addressing the limited expressive ability of the parts-to-wholes assignment in computer vision. In natural language processing community, however, the capsule network has not yet been widely investigated. \newcite{zhao2018investigating} testify capsule network on text classification and \newcite{gong2018information} propose to aggregate a sequence of vectors via dynamic routing for sequence encoding. \newcite{dou2019dynamic} first propose to employ capsule network in NMT using routing-by-agreement mechanism for layer representation aggregation. \newcite{wang2019towards} develops a constant time NMT model using capsule networks.
These studies mainly use capsule network for information aggregation, where the capsules could have a less interpretable meaning. In contrast, our model learns what we expect by the aid of auxiliary learning signals, which endows our model with better interpretability.

\section{Conclusion}
In this paper, we propose to recognize the translated \past\ and untranslated \future\ contents via parts-to-wholes assignment in neural machine translation. We propose the guided dynamic routing, a novel mechanism that explicitly separates source words into \past\ and \future\, guided by \present\ target decoding status at each decoding step. We empirically demonstrate that such explicit separation of source contents benefit neural machine translation with considerable and consistent improvements on three language pairs. Extensive analysis shows that our approach learns to model the \past~and \future\ as expected, and alleviates the inadequate translation problem. It is interesting to apply our approach to other sequence-to-sequence tasks, e.g., text summarization (as listed in Appendix).

\section*{Acknowledgement}
We would like to thank the anonymous reviewers for their insightful comments. Shujian Huang is the corresponding author. This work is supported by the National Science Foundation of China (No. U1836221 and No. 61772261), the Jiangsu Provincial Research Foundation for Basic Research (No. BK20170074).

\bibliography{iclr2019_conference}
\bibliographystyle{acl_natbib}

\clearpage

\appendix

\section{Machine Translation}

\subsection{Detailed Experimental Settings}
For Zh-En, we segmented Chinese data using ICTCLAS\footnote{\url{http://ictclas.nlpir.org/}}. We limited the maximum sentence length to 50 tokens. For En-De and En-Ro, we did not filter out the sentence length for En-De and En-Ro. We applied byte pair encoding~\cite[BPE]{Sennrich2016Neural} to segment all sentences with merge operations of 32K. All out-of-vocabulary words were mapped to a distinct token \texttt{<UNK>}.  

We used the Adam optimizer~\cite{KingmaB14:adam:iclr} with $\beta_1=0.9$, $\beta_2=0.98$, and $\epsilon=10^{-9}$. We used the same learning rate schedule strategy as \cite{Vaswani2017Attention} with 4,000 warmup steps. The training batch consisted of approximately 25,000 source tokens and 25,000 source and target tokens. Label smoothing of the value of  0.1~\cite{szegedy2016rethinking} was used for training. We trained our models for 100k steps on single GTX 1080ti GPU. 

For evaluation, we used beam search with a width of 4 with length penalty of 0.6~\cite{wu2016google}. We did not apply checkpoint averaging~\cite{Vaswani2017Attention} on the parameters for evaluation. The translation evaluation metric is case-insensitive BLEU \cite{papineni2002bleu} for Zh-En\footnote{\url{https://github.com/moses-smt/mosesdecoder/blob/master/scripts/generic/multi-bleu.perl}}, and case-sensitive BLEU for En-De and En-Ro\footnote{\url{https://github.com/awslabs/sockeye/tree/master/contrib/sacrebleu}}, which are consistent with previous work.


\begin{figure}[t] 
  \centering
  \subfloat[BLEU scores regarding dimensionality of each capsule.] {
    \label{fig:hyper1}
    \includegraphics[width=0.38\textwidth]{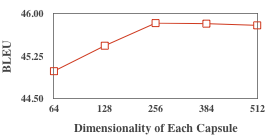}
  } \\
  \subfloat[BLEU scores regarding numbers of capsules of each category.] {
    \label{fig:hyper2}
    \includegraphics[width=0.38\textwidth]{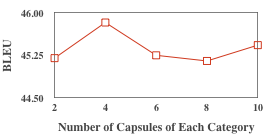}
  } \\
  \subfloat[BLEU scores regarding routing iterations of {\sc Gdr}] {
    \label{fig:hyper3}
    \includegraphics[width=0.38\textwidth]{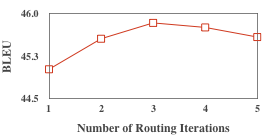}
  } 
  \caption{BLEU scores in terms of different hyperparameters.}
  \label{fig:hyper}
\end{figure}

\subsection{Effect of Hyperparameters} 
To examine the effects of different hyperparameters of the proposed model, we list the results of different settings of the dimension and the number of the capsules, and the number of routing iteration in Table \ref{fig:hyper}.

We observe that increasing the dimension or the number of capsules does not bring better performance. We attribute these results to the sufficient expressive capacity of medium scales of the capsules. 
Likewise, the BLEU score goes up with the increase of the number of iterations, while it turns to decrease after the performance climb to the peak at the best setting of 3 iterations. The number of routing iteration affects the estimation of the agreement between two capsules. Hence, redundant iterations may lead to over-estimation of the agreement, which has also been revealed in \newcite{dou2019dynamic}. 

\begin{CJK}{UTF8}{gbsn}
\begin{figure*}[t] 
  \centering
  \subfloat[An example of guided dynamic routing. The orderings of the source sentence in Chinese and the English translation are non-monotonic. For example, after the target word ``supply'' has been generated in the intermediate of the translations, the assignment probabilities of its corresponding source word ``供给'', near the end of the source sentence, changes from the \past\ to the \future. 
  ] {
    \label{fig:vis2_1}
    \includegraphics[width=0.95\textwidth]{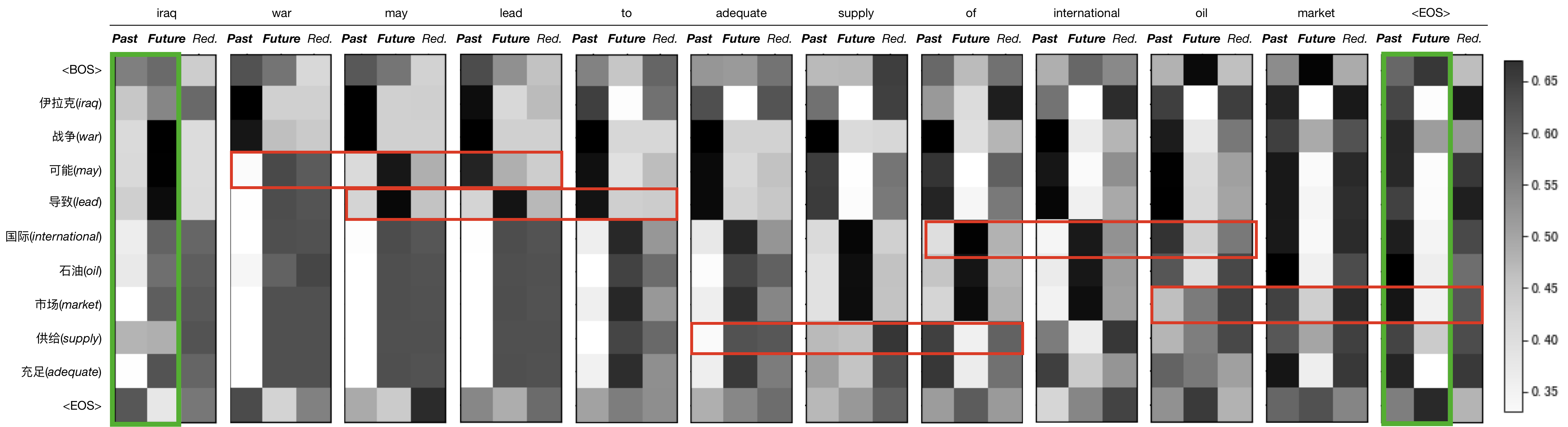}
  } 
  
  \subfloat[Another example. An interesting case is related to ``unware'', which has a source counterpart of three Chinese words (``毫无'', ``所'', and ``知''). When it has been generated, the assignment probabilities of the words in its counterpart phrase change from \past\ to the \future simultaneously. ] {
    \label{fig:vis2_2}
    \includegraphics[width=0.98\textwidth]{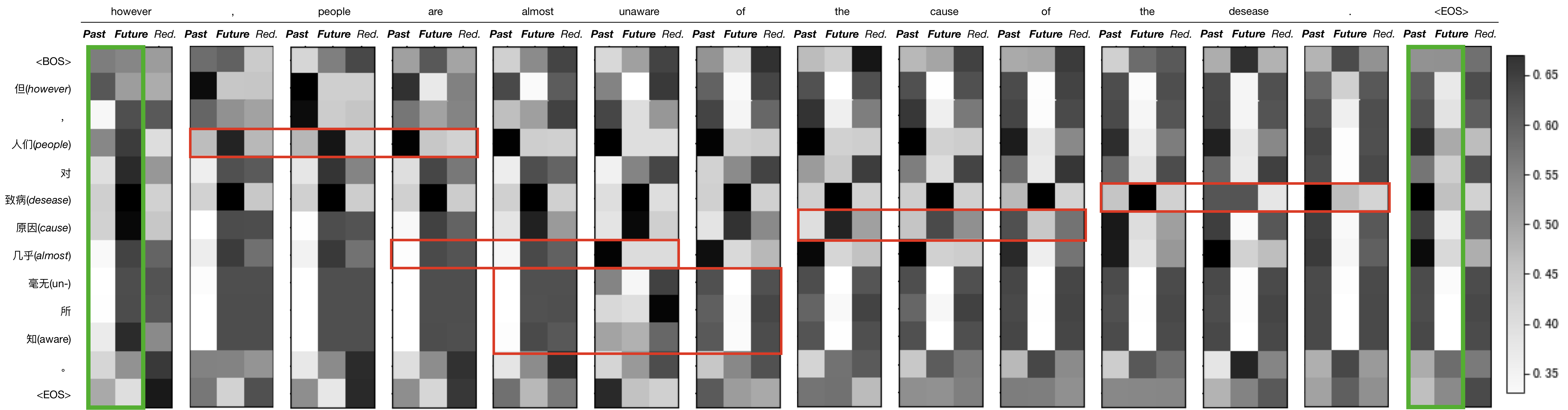}
  }
  \caption{Visualization of the assignment probabilities of the iterative guided dynamic routing. For each translation word, the three columns represent the probability assigning to \past, \future\ or redundant capsule, respectively. Green frames indicate that the assignment probabilities of source words change from \past\ at the beginning to \future\ in the end. Red frames highlight the changes before and after a specific word's generation. }
  \label{fig:case-visualization2}
\end{figure*}
\end{CJK}

\subsection{Visualization}
We show two more examples of visualization of the assignment probabilities of the guided dynamic routing mechanism in Figure \ref{fig:case-visualization2}.

\section{Abstract Summarization}
Abstract summarization is another prevalent sequence-to-sequence task, which aims to find a short, fluent and effective summary from a long text article~\cite{chang2018ahw}. Intuitively, discarding redundant contents in the source article is important for abstract summarization, which could be achieved by our proposed redundant capsules. Moreover, figuring out which parts of the source article have been summarized completely and which have not yet should be also explicitly modeled for abstract summarization.  

\subsection{Experimental Settings}
\paragraph{Dataset} We conduct experiments on the LCSTS dataset~\cite{Hu2015LCSTSAL} to evaluate our proposed model for abstract summarization. This dataset contains a large number of short Chinese news articles with their headlines as the short summaries collected from Sina Weibo, a Twitter-like micro-blogging website in China. As shown in Table \ref{tab:dataset-lcsts}, this dataset is composed of three parts. Part I contains a large number of 2,400,591 pairs of (article, summary). Part II and III contain not only text data but also manually rated scores from 1 to 5 for the quality of summaries in terms of their relevance to the source articles. We follow~\newcite{Hu2015LCSTSAL} to use Part I as the training set, and the subset scored from 3 to 5 of Part II and Part III as the development set and testing set.

\begin{table}[h]
  \footnotesize
  \centering
  \begin{tabular}{l|ccc}
    \toprule
            & Part I    & Part II   & Part III      \\
    \hline
    $\#$pairs           & 2,400,591 & 10,666    & 1,106  \\
    $\#$pairs (score $\geq 3$ )           & - & 8,685    & 725  \\
    \hline
    \end{tabular}
  \caption{Statistics of the LCSTS dataset.}
  \label{tab:dataset-lcsts}
\end{table}

\paragraph{Training and evaluation} Following \newcite{Hu2015LCSTSAL}, we conducted experiments based on character. We set the vocabulary size to 4,000 for both source and target sides. We used the Transformer as our baseline system. Model configurations and other training hyperparameters are the same as machine translation tasks. For evaluation, we used beam search with a width of 10 without length penalty. We report three variants of recall-based ROUGE~\cite{Lin2004ROUGEAP}, namely, ROUGE-1 (unigrams), ROUGE-2 (bigrams), and ROUGE-L (longest-common substring). 

\subsection{Results}

\begin{table}[h]
  \footnotesize
  \centering
  \begin{tabular}{l|ccc}
    \toprule
    \bf Model               & R-1       & R-2       & R-L      \\
    \hline
    RNN+context~\cite{Hu2015LCSTSAL}            & 29.9  & 17.4  & 27.2  \\
    CopyNet~\cite{P16-1154}                     & 34.4  & 21.6  & 31.3  \\
    Distraction~\cite{chen2016distraction}      & 35.2  & 22.6  & 32.5 \\
    DGRD~\cite{li2017deep}                      & 36.99 & 24.15 & 34.21  \\
    MRT~\cite{ayana2016mrt}                     & 37.87 & 25.43 & 35.33 \\
    WEAN~\cite{ma2018query}                     & 37.80 & 25.60 & 35.20  \\
    AC-ABS~\cite{li2018actor}                   & 37.51 & 24.68 & 35.02 \\
    Transformer~\cite{chang2018ahw}             & 40.49 & 26.83 & 37.32  \\
    ~~~~+HWC~\cite{chang2018ahw}         & 44.38 & 32.26 & 41.35 \\
    \hline
    Transformer             & 40.18 & 25.76    & 35.69  \\
    \textsc{Ours}           & \bf 43.85 & \bf 29.57    & \bf 39.10 \\
    ~~~~- \textit{redundant capsules}   & 42.43 & 28.17    & 37.66  \\
    \hline
    \end{tabular}
  \caption{ROUGE scores on LCSTS abstract summarization task.}
  \label{tab:exp-lcsts}
\end{table}

Table \ref{tab:exp-lcsts} shows the results of existing systems and our proposed model. We observe that our proposed model outperforms the Transformer baseline system by a significant margin. As we expected, the redundant capsules are vital for abstract summarization task, in which summary is required to be short and concise by discarding tons of less important contents from the original article. Our model also beats most of the previous approaches except the model of~\newcite{chang2018ahw}, which benefit from a hybrid word-character vocabularies of max size (more than 900k entries) and data cleaning. We expect our model to benefit from these improvements as well. In addition, our model does not rely on any extraction-based methods, which are designed for abstract summarization task to extract relevant parts from the source article directly (e.g., CopyNet~\cite{P16-1154}). Our method may achieve further gains by incorporating these task-specific mechanisms.

\end{document}

%% file: math_commands.tex
\usepackage{amsmath,amsfonts,bm}

\def\Figref#1{Figure~\ref{#1}}

\def\eqref#1{equation~\ref{#1}}

\def\1{\bm{1}}

\def\rvh{{\mathbf{h}}}

\def\vtheta{{\bm{\theta}}}
\def\va{{\bm{a}}}

\def\vh{{\bm{h}}}

\def\vo{{\bm{o}}}

\def\vs{{\bm{s}}}

\def\vv{{\bm{v}}}
\def\vw{{\bm{w}}}

\def\vz{{\bm{z}}}

\def\mE{{\bm{E}}}

\def\mW{{\bm{W}}}

\DeclareMathAlphabet{\mathsfit}{\encodingdefault}{\sfdefault}{m}{sl}
\SetMathAlphabet{\mathsfit}{bold}{\encodingdefault}{\sfdefault}{bx}{n}

